\documentclass[pdflatex,sn-apa]{sn-jnl}% Math and Physical Sciences Reference Style
\usepackage[T1]{fontenc}
\usepackage{booktabs}
\usepackage{pdflscape}
\usepackage{ragged2e}
\usepackage{array}
\jyear{2021}%

\theoremstyle{thmstyleone}%
\theoremstyle{thmstyletwo}%

\theoremstyle{thmstylethree}%

\begin{document}

\title[Models of symbol emergence in communication]{Models of symbol emergence in communication: a conceptual review and a guide for avoiding local minima}


\author*[1]{\fnm{Julian} \sur{Zubek}}\email{j.zubek@uw.edu.pl}

\author[2]{\fnm{Tomasz} \sur{Korbak}}

\author[1]{\fnm{Joanna} \sur{Rączaszek-Leonardi}}

\affil*[1]{\orgdiv{Faculty of Psychology}, \orgname{University of Warsaw}, \orgaddress{
\street{Stawki 5/7}, \city{Warsaw}, \postcode{00-183},
\country{Poland}}}

\affil[2]{\orgdiv{Department of Informatics}, \orgname{University of Sussex}, \orgaddress{\country{UK}}}


\abstract{Computational simulations are a popular method for testing hypotheses about the emergence of communication. This kind of research is performed in a variety of traditions including language evolution, developmental psychology, cognitive science, machine learning, robotics, etc. The motivations for the models are different, but the operationalizations and methods used are often similar. We identify the assumptions and explanatory targets of several most representative models and summarise the known results. We claim that some of the assumptions – such as portraying meaning in terms of mapping, focusing on the descriptive function of communication, modelling signals with amodal tokens – may hinder the success of modelling. Relaxing these assumptions and foregrounding the interactions of embodied and situated agents allows one to systematise the multiplicity of pressures under which symbolic systems evolve. In line with this perspective, we sketch the road towards modelling the emergence of meaningful symbolic communication, where symbols are simultaneously grounded in action and perception and form an abstract system.}




\maketitle

\section{Introduction}

The problem of the emergence of communication systems lies at the heart of theories of cognition. It subsumes the problem of the emergence of symbols that seem conceptually indispensable for formulating explanations in cognitive science and artificial intelligence. Over the last two decades, this problem has been tackled using computational models, which made the theories more internally coherent and, at the same time, more comparable and classifiable. Models are grounded in different domains of inquiry – from philosophy to linguistics, to robotics – and motivated by questions relevant to those domains; yet, despite this variety, they seem to share some basic assumptions about the process of communication. We attribute these assumptions to two different sources.

The first source is the view of communication imposed by the influential work of Shannon in the mid-last century and strengthened by the facilitation of computational modelling that such a perspective carries. In most models, communication is understood as conveying messages from an information source to a destination via signals transmitted over a communication channel \citep{shannon1948}. In Shannon's own words: ``The fundamental problem of communication is that of reproducing at one point either exactly or approximately a message selected at another point''. This highlights the problem of message transmission as central for communication. The “meaning” of messages for a long time was in the second plane, usually simplified as a mapping between symbols and things in the agent environment, which realises the “grounding” of symbols \citep{harnad1990}. The message-object mapping, representing meaning, mirrors the message-code mapping central to information theory.

However, the fathers of information theory were convinced that the term communication should have a broader meaning. As \cite{weaver1949} put it: “The word communication will be used here in a very broad sense to include all of the procedures by which one mind may affect another”. This resonates with more pragmatic accounts of meaning that focus on influencing agent actions rather than representing the world. In human communication, this influence is often based on existing conventions. For example, I can signal a person to come closer or stay away using conventional gestures. This brings us to the second source of assumptions concerning communication: Lewis signaling game (from now on LSG; \cite{lewis1969}). On the ground of game theory, Lewis demonstrated that a conventional communication protocol emerges as an equilibrium when two agents -- sender and receiver -- are rewarded for successful coordination; it is rational for the two agents to stick to the convention, even if the convention itself is arbitrary. This was Lewis’ solution to the problem of meaning creation by convention. Notably, in the original formulation of LSG information-theoretical tools were not used, and Shannon's communication theory was not cited as a reference. Later, the two approaches were integrated by \cite{skyrms2010}. Subsuming LSG into the general information-theoretical schema for communication, he was able to interpret signals as probabilistic events that carry information and conceptually integrate information quantity and information content (conventional meaning). However, the consequence of this move was undermining some of the pragmatic inclinations of the original Lewis work. Skyrms noted that in many situations signaling can be discussed equivalently in terms of information flow and in terms of coordination, but he preferred the first perspective, which reversed the original emphasis of Lewis.\footnote{In \cite{lewis1969} the word “information” appears 11 times, and the word “coordination” 138 times, in \cite{skyrms2010} the word “information” appears 358 times, and the word “coordination” 7 times.} This made it easy to once more conflate signaling with information transmission and to see LSG as part of the meaning-mapping enterprise.

The resulting signaling game framework, realised by many computational models, seems to be compatible with the concept of language games, pioneered by Wittgenstein (\cite{wittgenstein1953}, see also \cite{correia2020}). A classical example of a language game would be a builder who uses building stones of various shapes and asks his assistant to pass a stone of a particular shape. The language game results in success when the right kind of stone is passed. These kinds of structured interactions are simulated with artificial agents, the behaviour of which is governed by some internal adaptive mechanisms (for example, neural networks). At first, agents do not have any meaningful predispositions (policy), so their actions and signals that they produce are accidental. Each episode of the game brings consequences for the agents (for instance, success or failure) determining their reward, which make them update their predispositions. Through individual adaptations, agents become more successful, but at the same time, they change the rules of the game for other agents. For example, if an agent adopts the word “glud” to refer to a cuboid-shaped stone, other agents need to pick up on that to successfully interact with the first agent. In this aspect, the adaptation process resembles coevolution within a population, rather than a simple optimisation of individual fitness. Computational simulations demonstrate how this iterative adaptation can lead to the emergence of stable communication systems.

Existing work on emergent communication has repeatedly demonstrated the possibility of learning a conventionalised mapping from a set of discrete signals to a set of objects without any central authority. Optimal mapping can be learnt under various settings, agent architectures, and adaptation mechanisms \citep{spike2017}. Natural languages, however, with their richness and expressiveness, do not resemble simple bidirectional mappings. They include parts of speech that are not easily mapped to objects or events, they involve complex syntax and grammar, they tolerate nuances and ambiguities (and exploit them, as in the case of poetry), they allow speculative conversations on abstract entities, and they crucially depend on interaction in time. Much work is still needed to build an adequate account of these phenomena within the field of language emergence modelling. But the direction in which to proceed is unclear. In this paper, we claim that the field is in a state of a local minimum created by a particular set of assumptions. This obscures the view, taking some properties of the communication process for granted and leaving crucial aspects of language unexplored.

Getting stuck in the above-mentioned local minimum hinders the efforts to scale up from simple signaling games to structured communication systems, which can be used as complex compositional controls in time-dependent interactions within structured environments. Recent developments in cognitive science, linguistics, psycholinguistics, and linguistic anthropology point to multiple processes and properties of communication, which appear to be overlooked in dominant models and might be crucial for this kind of scaling up. Theories that ground communication in biological information processes \citep{pattee1969, pattee1982, deacon2011}, interaction-dominant and pragmatics-oriented processes \citep{kempson2000}, dynamical systems models of cognition \citep{raczaszek-leonardi2008} and enaction \citep{dipaolo2018} offer a richer perspective on the nature and functions of communication. Bringing them into play on the modelling level may be helpful in overcoming some of the obstacles.

In this paper, we review several attempts of emergent communication modelling that we deem representative to the multitude of models that have appeared over the last two decades and evaluate them in terms of their merits and shortcomings. This will allow a better grasp of their underlying common assumptions about the process of communication and identification of those, which might be responsible for hindering the progress towards modelling the emergence of structured symbolic systems. The main claim of the paper is that progress does not hinge on the complexification of the existing models, making them more efficient computationally or applicable to more complex environments, nor even in a creative composition of the models. We claim that in order to leave the local minimum, we need to examine closely and perhaps suspend some of the core, and very basic, assumptions on which the extant modelling efforts rest. For this, we perhaps need to back out of some habits of thinking and seek guidance in those theories which do not assume an easy reduction of the problem of communication to the engineering problem of information transmission. From this widened perspective, we can recognise that language is not a product of a single optimisation process, but is shaped by multiple factors. Thus, sources of complexity of language are multiple and may possibly interact, leading to emergent properties. It is possible that language properties that have simple formal description, for example recursively nested structures described with formal grammars, could nevertheless be shaped by multiple interacting pressures. Opening up this possibility will hopefully set novel explanatory targets for the models and open new avenues for research on emergent symbolic communication.

\section{Modelling traditions and explanatory targets}

\begin{figure}[h]%
\centering
\includegraphics[width=0.8\textwidth]{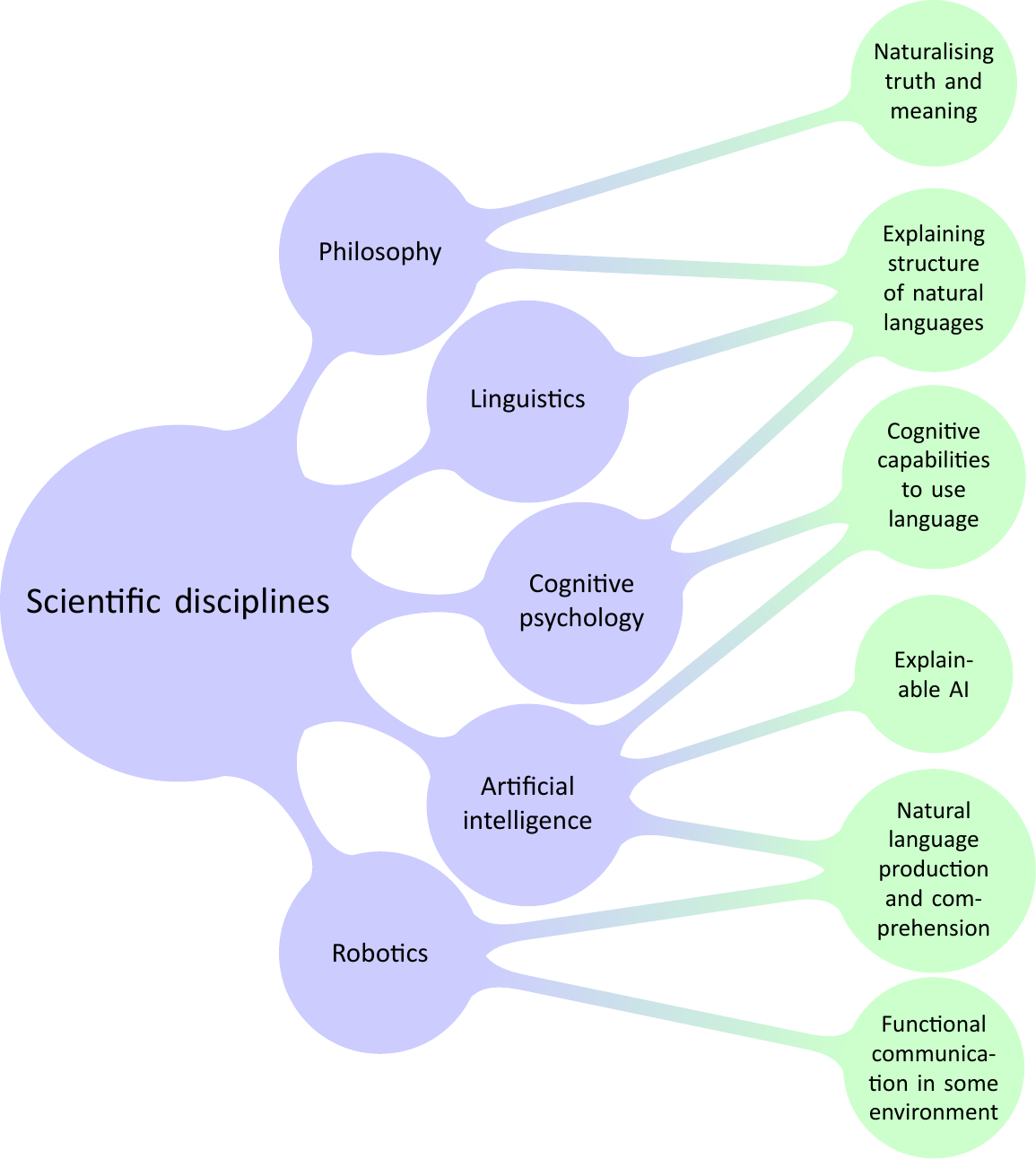}
\caption{Different scientific disciplines and the roles they assign to emergent communication models.}\label{fig1}
\end{figure}

Computational models of the emergence of communication come from a variety of research traditions, reflecting their theoretical goals, definitions, and methods. Giving justice to their recent proliferation is beyond any single publication. We roughly group them by discipline to draw the reader's attention to different explanatory roles assigned to the simulations. These arise from differences in the perspectives on language in general and on the place, role, and nature of language within human cognition and behaviour. We will point out how those differences are manifested in computational modelling and use the presence and prominence of such models in a discipline as a criterion to include them in this review. An (admittedly simplified) picture in Figure~\ref{fig1} illustrates some of the commitments and explanatory targets of the models in different scientific disciplines (the division of the domains is an idealisation). 

Philosophical reflection on the nature of language and its relation to thought is probably as old as philosophy itself. In this context, game-theoretic models have been used in research of social conventions, normativity, and semantic information, starting from the influential work of \cite{lewis1969}. He formalised the notion of convention, which arises in the interaction of agents. The sender has to communicate the state of the world to the receiver by emitting discrete signals, upon which the receiver produces an adequate reaction. Lewis' original motivation was to provide a game-theoretic explication for the notion of truth by convention (and defend it against \cite{quine1936}). His ideas influenced many other authors.

When implemented as computer simulations, these models are kept abstract and usually consist of disembodied agents passing discrete tokens. The goal is not to faithfully portray actual processes but rather to support certain philosophical positions by demonstrating the theoretical possibility of the bottom-up emergence of conventions. In recent years, this thread has been developed to account for more complex communication processes \citep{skyrms2010, barrett2017, barrett2019, korbak2021a}.

In the area of linguistics, computational models of language emergence are used to ask a variety of questions. Contrary to biological evolution, language as such (before writing) leaves no “fossils”, so the communicative systems of early humans (or hominids) are difficult to uncover. In this context, agent-based simulations became attractive tools for testing hypotheses about the origins of language and possible evolutionary trajectories, both on the biological and cultural time scales. The goals are theoretical: models concern various aspects of language, such as evolution of lexemes and systematic variation of syntax (\cite{nowak1999}; for a review, see, e.g., \cite{ferri2018, grifoni2016}) or evolution of vocabulary. Compared to the philosophical tradition, simulations developed within evolutionary linguistics tend to be more focused on language structure. The results of the simulations are compared with the empirical data to assess their plausibility as models of language evolution and change \citep{brighton2005, parisi2008, deboer2010}.

In the field of cognitive psychology, the emphasis is on the connection between communication and language and other cognitive processes, mainly internal, such as memory, categorisation, and reasoning. For example, emergent language, stabilised in language games, has been shown to be a factor determining the shared structure of individual categories within populations \citep{steels2005, baronchelli2015}. Some models in this field do concern the emergence of language on the developmental timescale, although, curiously and symptomatically, these rarely take the form of computational simulation. However, an interesting turn can be noticed from the focus on learning as tuning of the predetermined generative structure, resulting in the production of a well-formed language \citep{chomsky1957}, to a focus on how developmental and functional processes might be reflected in the language structure \citep{ deacon1997, brighton2005}. In the former approach, which spurned modern psycholinguistics, cognitive psychology was seen as a subsidiary to linguistics, responsible for less theoretically noble processes of realisation (performance; \cite{devitt2003}), while in the latter it comes to the fore as a source of processes that structure language \citep{raczaszek-leonardi2009}. Some models address directly how language structure is shaped by and, in turn, shapes other processes (cognitive, environmental, or social). For example, models explored the effects of ``starting small'', that is, with a smaller memory capacity \citep{elman1993}, or of interactive factors within social environments or various demographic characteristics such as population size or social structure on language evolution \citep{vogt2009, dale2012}.

Emergent communication models have also enjoyed a recent surge of interest in the machine learning and artificial intelligence (AI) community. Consistently with the original mission of AI, here research is focused on replicating cognitive processes which are necessary to develop and use human intelligence, in this case language.  A great deal of machine learning research on emergent communication addresses fundamental questions concerning the structure of language. \cite{lazaridou2018a} demonstrated that reinforcement learning-driven agents are capable of solving the LSG in which input is given as raw pixel images, not symbolic tokens. Emerged protocols displayed some degree of systematic structure, although ad-hoc vocabulary fitted to a particular game was preferred. Several other studies focused on the circumstances under which a more structured language appears \citep{mordatch2018, andreas2019, chaabouni2020}. Other models are not motivated directly by replicating human cognitive processes but arther by engineering problems related to natural language understanding. \cite{kottur2017a} work on emergent dialogue is to some extent motivated by the work on visually-grounded conversational systems \citep{mostafazadeh2017}. The formulation of the problem in \cite{lazaridou2017} is motivated by image captioning \citep{karpathy2017}. \cite{lee2018} posed the problem of unsupervised visually grounded translation as a LSG. The potential applications of emergent communication research in explainable AI \citep{doshi-velez2017} are also an intense area of research. They focus mainly on translating a communication protocol emerging between agents in the context of some task into a natural language such as English. For visually grounded approaches, this can be done using labelled images by relying on visual semantics determined by the class hierarchy of the images \citep{lazaridou2017}. Another approach is to align utterances in English and in an emergent communication protocol with respect to the effects they exert on a trained receiver. This also paves the way for natural language understanding by artificial agents.

The trend of making simulations more realistic and embodied is taken even further in artificial life and cognitive robotics communities. In this tradition, agents are not abstract entities but autonomous mobile robots (usually simulated in silico to some extent). Their communication is intertwined with the rest of their activity, often involving exploration of the environment and search for food or other agents \citep{floreano2007, grouchy2016, schulz2012}. Stress is placed primarily on the robustness of emerging communication protocols and minimal starting assumptions, as communication should scale up to realistic environments. More advanced communication features are built from the bottom-up rather than arbitrarily imposed. Some works in this tradition aim to explain basic aspects of animal communication in different evolutionary scenarios \citep{floreano2007, mitri2009}, while others target phenomena characteristic for human languages, such as its productivity, presence of grammatical structures, and the focal role of dialogue \citep{grouchy2016, schulz2012}. In contrast to other domains, signals appearing in robotics models are often modulated by multiple factors and are not fully controlled by agents. For example, in \cite{floreano2007} communication is realised through light signals that have colour regulated by the sender, but also properties dependent on the physics of the environment (directionality, possibility of being obscured by obstacles, etc.).

As can be glimpsed from the above, each discipline, which includes communication in its area of interest, has explanatory targets dependent on its general nature and goals. Models are designed to address various aspects of communication and language emergence, sometimes using different techniques. It seems that one of the problems in constructing more comprehensive models of language emergence is the compartmentalization of scientific efforts into different domains. The domain in which we see the greatest proliferation of models and in which most novel techniques appear is AI, where research is often motivated by concrete engineering problems and not theory of language per se. The novel modelling techniques do radiate to the basic problems and their respective domains, however, already tinted by the specific approach. On the other hand, recent advances in the respective fields that are interested in the theory of language emergence (developmental linguistics, language evolution, enactivism) are rather slow to seep into the computational models. In consequence, designed models often rely on a simplified treatment of the basic problems, which are the core of more theoretically rich approaches, and modelling results cannot be easily incorporated into constructed theories. Here, cognitive science, integrating multiple disciplines from linguistics to psychology to anthropology to computer science, has a good chance to serve as a bridge between the domains. Importantly, due to the recent “ecological turn” toward action-oriented and situated cognitive science, it has the opportunity to integrate processes over the biological, developmental and communication timescales \citep{raczaszek-leonardi2003, christiansen2022}.

We identify ourselves with the positions of such ecological cognitive science. From this perspective, we point out how this lack of flow of ideas between domains might slow down progress of computational modelling as a significant aid for the theoretical models of communication and language emergence. In the following, we describe concrete examples of models which best illustrate the various theoretical goals. We will show their advantages and contributions, but also indicate the possible simplifications, which will be the basis for conclusions regarding further progress. 

\section{From Lewis signaling game to embodied simulations}

Despite different distributions of accents of the above domains, recently they have increasingly used computational simulations implemented as agent-based models with explicit agent architectures. This facilitates the recognition of basic assumptions regarding the communication and language on which they are based. As a blueprint for the analysis of all kinds of modelling efforts, we adopt Lewis signaling game (LSG). As a minimal architecture for an interactive communicative setting, it delimits a broad class of models, capturing the essential elements of the problem of convention- or signal-based social coordination. It is recognised and sometimes explicitly built-on across various disciplines (from philosophy and anthropology to psychology, biology, economics, etc.). This should further facilitate the comparison of the models on a conceptual level, in terms of modifications of this basic scheme for various reasons, which can be evaluated as welcome or unwelcome from the theoretical point of view. Showing their basic compatibility with the LSG should also reveal that they share limitations inherent in the blueprint itself, if indeed such limitations are present.

LSG consists of two agents, a sender and a receiver, a set of states of affairs (stimuli), a set of signals, and a set of responses. A sender observes a state of nature, sends a signal to the receiver, and the receiver chooses a response. There is a unique correct response given the state of affairs unknown to the receiver. If the receiver chooses the correct response, both the sender and the receiver are rewarded. In the course of learning (or evolution), the agents converge to a Nash equilibrium such that the receiver always chooses the correct response. This means that the agents must have agreed upon a convention: a mapping from states of nature to signals and from signals to responses. Lewis called this a \emph{signaling system} (nowadays it is also called a communication protocol).

The original interpretation of the LSG involved rational agents maximising their reward. \cite{skyrms2010} later reinterpreted it in terms of evolutionary game theory: he demonstrated that signaling systems may emerge through simple replicator dynamics (which may represent both biological and cultural evolution). He proposed that the information present in the signaling system is sufficient to provide a general account of semantic meaning. According to Skyrms, meaning emerges in the symmetry breaking of converging to one signaling system (out of several possible). Each signal can be assigned a descriptive meaning (when an external observer estimates the probability of the sender perceiving certain stimulus given the signal) and indicative meaning (when an external observer estimates the probability of the receiver performing certain action given the signal). Such meaning is a property of the signal system as a whole and does not presuppose that the sender and receiver are rational agents introducing any kind of meaning of their own.

Several other problems in philosophy and social science can be formalised in terms of LSGs. The labour market can be modelled in terms of employees (senders) who signal their ability to employers (receivers) through educational credentials (signals) \citep{spence1973}. Voter ignorance and the spread of misinformation can be modelled in terms of people (senders) maintaining or spreading false beliefs (signals) to signal their group identity or political affiliation \citep{funkhouser2017}. \cite{barrett2017} recently considered generalised LSGs consisting of multiple subgames. These include games that evolved from simpler games through template transfer or modular composition. Template transfer assumes that some adaptations that emerged in the context of one signaling game can be transferred to another game. One of the possibilities would be to reuse the same structure of emitted signals and adapt them for a new set of stimuli. Template transfer assumes the historicity of the evolution or learning process giving rise to the signaling systems – previous adaptations pave the way for the new ones. The games themselves are independent and, in principle, could also be learnt in another order or in parallel. In contrast, \emph{modular composition} is based on recursively nested games, where a simpler game becomes a part of a more complex one. A variety of social, cognitive, and semiotic phenomena can be understood in terms of generalised signaling games, including logical inference \citep{barrett2017, lacroix2022}, information integration in a network of agents \citep{barrett2019}, functional specialisation, and communication structuring \citep{barrett2018}.

\subsection{Beyond Lewis signaling game}

Most of the recent models of emergent communication may be seen as extensions of the LSG with respect to particular elements: stimuli, signals, responses, or reward/learning mechanism. Below, we provide an overview of various models, from the simplest extensions to complex embodied models, which diverge from the original structure of LSG significantly. We do not mean to provide a comprehensive review. Our choice of models is dictated by their popularity and representativeness for different strands of modelling efforts. In our overview, we discuss how these models aim to explain different sources of communication complexity, connected to a) internal structure of signals, b) structure of the environment, c) structure of actions in the environment and types of control introduced by signals, and d) structure of rewards.  The selected models and their properties are presented in Table 1. In the next sections, we discuss them in detail starting from relatively simple formal models and moving towards embodied simulations in realistic environments. Throughout, we refer to the LSG as a blueprint and demonstrate how the presented models deviate from its structure (as reflected by section titles).

\begin{landscape}
\begin{table}[!htp]\centering
\caption{Discussed models of emergent communication from various traditions and their characteristics in terms of the basic template of LSG. The models were selected in such a way as to display various interaction scenarios considered as communicative situations in the model. We ordered them according to their abstractness: from idealised formal models to embodied simulations/mobile robots.}\label{tab:models}
\begin{scriptsize}
\begin{tabular}{p{2cm}p{1.2cm}*{5}{>{\RaggedRight}p{2.5cm}}}
Model &Discipline &Environment (stimuli) &Signals &Available actions &Rewards and learning mechanism &Results/explanatory target \\\midrule
\cite{lewis1969} &philosophy &Discrete tokens &Discrete tokens &Discrete tokens &Specific response rewarded &Stabilization of social conventions \\\midrule
\cite{nowak1999} &linguistics &Discrete tokens &Sets of tokens (compound expressions), risk of confusion &Discrete tokens &Different rewards dependent on the response &Evolutionary origins of phonological and grammatical structures \\\midrule
\cite{smith2003}  &linguistics &Binary vectors: combinations of properties &Sets of tokens (compound expressions) &- &Learning mapping after previous generation in supervised manner &Origins of systematic structure \\\midrule
\cite{batali1998} &linguistics &Binary vectors: simple predicates &Sequences of tokens (compound expressions) &Binary vectors: reconstructed predicates &Sender tries to predict receiver’s responses internally (obverter). Receiver learns from the sender in a supervised manner. Roles change periodically. &Emergence of systematic structure in sequences of variable length \\\midrule
\cite{jaques2019} &artificial intelligence &2D grid world with simple objects encoded as binary vectors &Discrete tokens &Movement in the environment &Number of collected food items rewarded, group reward &Emergence of communication under intrinsic social influence motivation \\\midrule
\end{tabular}
\end{scriptsize}

\end{table}

\setcounter{table}{0}
\begin{table}[!htp]\centering
\caption{Continued from the previous page.}\label{tab:models2}
\begin{scriptsize}
\begin{tabular}{p{2cm}p{1.2cm}*{5}{>{\RaggedRight}p{2.5cm}}}

Model &Discipline &Environment (stimuli) &Signals &Available actions &Rewards and learning mechanism &Results/explanatory target \\\midrule
\cite{steels2005} &cognitive psychology &Sets of numerical vectors: CIE LAB colors (available to sender and receiver), marked stimulus (available to sender) &Discrete tokens from infinite set &Choosing correct stimulus &Specific response rewarded, supervised learning of new word forms &Emerging communication stabilizing category formation \\\midrule
\cite{lazaridou2017} &artificial intelligence &Sets of numerical vectors: raw RGB images &Sets of tokens (compound expressions) &Choosing correct stimulus &Specific response rewarded &Signal system grounded in raw visual input \\\midrule
\cite{cangelosi1999, cangelosi2001} &linguistics \& philosophy &2D grid world with objects encoded as binary vectors (available to sender and receiver, receiver gets perceptual noise) &Sets of tokens (compound expressions) &Discrete movement in grid-like environment and discrete response tokens &Locating food and avoiding poison rewarded in evolution, speaker not rewarded &Symbolic mode of reference – transfer of symbol–symbol relations to new vocabularies \\\midrule
\cite{schulz2012} &robotics &2D realistic environment, own position in the environment (available to sender and receiver) &Sets of tokens (compound expressions) &Continuous movement in the environment (available to sender and receiver) &Meeting at the appointed location rewarded &Constructing abstract location concepts grounded in spatial experience \\\midrule
\end{tabular}
\end{scriptsize}
\end{table}

\setcounter{table}{0}
\begin{table}[!htp]\centering
\caption{Continued from the previous page.}\label{tab:models3}
\begin{scriptsize}
\begin{tabular}{p{2cm}p{1.2cm}*{5}{>{\RaggedRight}p{2.5cm}}}

Model &Discipline &Environment (stimuli) &Signals &Available actions &Rewards and learning mechanism &Results/explanatory target \\\midrule

\cite{loula2010} &linguistics \& philosophy &2D grid world, 0-1 presence of nearby objects of various classes (available to all agents) &Discrete tokens from a large set &Choice of high-level, predefined behavior: wandering, fleeing, etc. (available to all agents) &Hebbian learning of associations (classical conditioning) &Shift between different forms of signification within an ecological niche \\\midrule
\cite{mordatch2018} &artificial intelligence &2D environment, (shared by sender and receiver), marked target object and action (available to sender) &Sets of tokens (compound expressions) &Continuous movement in the environment &Explicit error feedback &Systematic signal structure grounded in action structure \\\midrule
\cite{floreano2007} &robotics &2D environment (shared by all agents) &Light emitted by physical robots &Continuous movement in the environment (available to all agents) &Locating food and avoiding poison rewarded in evolution, speaker not rewarded &Communication strategies evolving under different kin structures \\\midrule
\cite{grouchy2016} &robotics \& linguistics &2D environment (shared by all agents) &Continuous scalar &Continuous movement in the environment (available to all agents) &No explicit rewards, two agents producing offspring upon contact &Emergence of complex mode of signification with signal-signal relations \\
\bottomrule
\end{tabular}
\end{scriptsize}
\end{table}

\end{landscape}

\subsection{Compound signals and noise}

The simple structure of LSG can be easily extended by adopting compound signals consisting of multiple tokens. This route was taken by \cite{nowak1999} who simulated the evolution of meaningful words (mono- and polysyllabic) and two word expressions. They were interested in the dynamics of language evolution. They investigated whether compound expressions could be evolutionarily preferred to simple words using methods of evolutionary game theory. They constructed a fitness function related to communicative success and identified language structures that optimise this function. The function included factors such as perceptual noise in the signal space (risk of confusing similar syllables) and uneven payoffs of successfully communicating different concepts (for example, signals warning against dangers of different magnitude). The authors demonstrated that under certain circumstances smaller and partially ambiguous lexicons are more optimal, and thus there are evolutionary limits on the growth of vocabulary. Compound expressions formed according to grammar rules restricting the space of valid utterances helped to transcend those limits without degrading communicative success. Sources of complexity here were the need to distinguish noisy signals and uneven payoffs: they provided evolutionary pressures making compound signals desirable. All simulations were kept on the abstract level: they demonstrated what sort of languages are optimal but did not simulate the emergence of actual communication properties.

The work of \cite{nowak1999} is a valuable demonstration of the role of noise and the complex structure of tasks that lead to uneven rewards in shaping the language structure. These factors align well with the ecological perspective. However, this model focused solely on the problem of signal recognition (identifying received signals as such and such) and not signal interpretation. It explains the form of the language and factors important for its stability under evolutionary pressures, but not the structure of meaning and not different modes of signification.

\subsection{Compositional signals via vertical transmission}

The interaction between the sender and receiver in a standard LSG can be called a horizontal interaction. In contrast, \cite{kirby2014} develop iterated learning games, which involve vertical interactions between subsequent generations of agents developing new vocabulary. In these games there are no actions or receivers. Instead, the sender alone produces a set of signals for the set of stimuli presented. Signals are compound multisyllabic utterances, and stimuli are objects described through the composition of stable properties. The stimulus-signal pairs produced become a learning sample for the next sender, which produces a learning sample for yet another sender, etc. This framework was mathematically analysed using evolutionary game theory \citep{brighton2002} and implemented in computational simulations. The goal of iterated learning is to explain the systematic structure of natural languages.

\cite{smith2003} constructed a simulation in which they represented agents as simple neural networks consisting of two layers of binary neurons. The first layer represented binary object properties (each object either had a particular property or did not have it), the second layer represented the presence of specific components in the utterance (each utterance was an unordered set of components). The connections between neurons were reinforced on the basis of the object-utterance pairs present in the learning input, and then a set of utterances for new objects was produced. The authors introduced a transmission bottleneck between generations: agents from the new generation observed only a subset of utterances produced by agents of the previous generation. They demonstrated that under these circumstances compositional languages -- in which individual syllables in an utterance are systematically mapped to object properties -- emerge. Importantly, \cite{kirby2008} replicated the same schema in an experiment involving human participants obtaining similar results.

Transmission bottleneck introduces evolutionary pressure towards structurization of meaning. This is different from the compound expressions with protogrammar analysed by \cite{nowak1999}. In that work, grammar defined only the structure of an utterance, demanding that the word of a particular class appears at a particular position in a sentence, which was in principle independent of the word's meaning. The structure of the compound expressions in \cite{smith2003} was not restricted (there was no grammar), but the way they referred to the properties of objects was systematic: different parts of the expression referred to different properties (the principle of compositionality).

Models based on vertical language transmission can be seen as orthogonal to horizontal LSG. They do not include a coordination problem, but rather a memorisation problem: agents learn the signaling protocol by observing examples rather than through interaction. Vertical and horizontal approaches can be seen as complementary. Recently, vertical transmission based on iterated learning is being introduced to horizontal communication models based on LSG. It was implemented by periodically resetting the weights of the neural networks of some agents (clearing their memory; \cite{li2019, cogswell2020}) or by having the agents passively observe the communication protocols of the previous generation \citep{ren2020}.

\subsection{Social reasoning facilitating communication}

A different line of work looks for pressures towards structured communication in pragmatic social reasoning, which is involved in language generation and comprehension. \cite{batali1998} attempted to explain the origin of the grammars of natural languages. In his simulations, the basic sender-receiver structure of LSG was preserved, but the sender used its own language comprehension module to predict the receiver responses. This prediction guided language generation -- an approach dubbed obverter \citep{oliphant1997}. Population of agents learnt in a self-supervised manner: a pair of agents selected at random interacted in each round, the sender acted as a teacher for the receiver, and, importantly, the roles were periodically swapped. The emerged language had compositional properties similar to \cite{smith2003}: some parts of the signal sequence had systematic meanings. Furthermore, it was shown that the agents were able to generalise and communicate novel combinations of meanings not present in the training set.

The obverter mechanism makes use of the same signal-meaning mapping for language generation and comprehension. This guarantees that agents are internally consistent and use the language they learn in the receiver role when they assume the speaker role. It can be interpreted as a rudimentary form of social reasoning: an agent's own cognitive capabilities serve as the model of another agent. The symmetry of the roles, together with the selection of speaker-receiver pairs at random from a population of agents, effectively creates a transmission chain similar to the iterated learning setting \cite{smith2003}. This should constitute a pressure for a systematic communication protocol. The obverter mechanism was implemented in modern neural networks by \cite{choi2018a} and \cite{bogin2019}, who also observed an increase in compositionality.

The more complex mechanism of social reasoning was explored in \cite{jaques2019}. Their research belongs to the domain of artificial intelligence and multi-agent cooperation.  They introduced the concept of social influence as an auxiliary reward for agents that communicate to solve a sequential social dilemma. Social influence was measured using counterfactual reasoning: an agent considered alternative signals it could have sent and assessed their likely effect on the actions of another agent. In one variant of their simulations, actions of the other agent were predicted using a specialised internal model trained in a supervised manner. Unlike in \cite{batali1998}, the model of the other agent was separated from the main model that controlled the actions of the agent. The authors did not focus on the signal structure, but showed that supporting communication with counterfactual reasoning leads to faster convergence, higher cumulative rewards, and higher mutual information between signals and actions. Even if agents did not have an explicit communication channel, influence rewards incentivised them to develop an ostensive communication protocol by pointing out important locations by staying still at these locations. This behaviour emerged because agents from the beginning viewed their actions not only as means to achieve their immediate goals but also as means to influence others. The dual meaning of the action followed the dual structure of the reward function.

The results of the simulations involving aspects of social reasoning suggest that some aspects of communication might be very hard to model if we treat language emergence as an unconstrained optimisation problem without particular inductive biases. In other words, agents might require certain cognitive preadaptations for social cooperation in order to establish a nontrivial communication protocol.

\subsection{Bidirectional influences between perception and communication}

\cite{steels2005} addressed the famous symbol grounding problem \citep{harnad1990} by tracking relations between language, categorisation, and perception. They used a variant of a language game which they called a guessing game. Similarly to the standard LSG, it involved a sender and a receiver communicating the properties of the presented stimuli. The stimuli were more complex than in the vanilla LSG (vectors of real numbers representing colours), and they were always presented in a context consisting of other stimuli. The sender was given a set of four random stimuli drawn from the continuous space with one distinguished target stimulus (the only restriction was that the target had to be sufficiently distinct from the other stimuli). Unlike the standard LSG, the receiver was presented with both the sender signal and the same set of stimuli. The task was to correctly choose the target stimulus. Each agent used a lexicon module -- a mapping between discrete categories and words -- and an adaptive categorisation module responsible for forming categories in a continuous perceptual space. Successful communication required coordination of both vocabulary and perceptual categories. In the simulation, a population of agents interacted randomly with each other freely exchanging sender and receiver roles. No colour categorisation was designed as “correct”. The authors demonstrated how a conventionalised categorisation of the colour stimuli may emerge through linguistic interaction. The vocabulary was also shared within the population and -- through the categorisation module -- grounded in the perceptual properties of the stimuli.

Similar models were used to study how aspects such as innate agents’ biases \citep{baronchelli2015} or the topology of a social network \citep{centola2015, zubek2017a} affect the formation of categories. However, all these works were interested in the structure of categories rather than the structure of language; the signals used in these simulations were simple unstructured tokens.

\subsection{Signals grounded in raw sensory inputs}

The path to ground signals in more realistic visual stimuli was explored further using deep neural networks. This kind of stimuli is drastically different from discrete tokens of the basic LSG. The departure from pre-processed symbolic input (e.g., encoding an object as a binary vector) in favour of raw visual input (i.e., arrays of RGB pixels) requires the sender to deal with entangled input and decompose it into relevant factors of variation (e.g. shape and colour) on its own – essentially performing image classification \citep{lazaridou2017, lazaridou2018a}. When compound signals are used, emergent communication protocols are on average more systematic with visual input compared to symbolic input. This corroborates the hypothesis that a structured environment plays a role in structuring language. Interestingly,  communication protocols do not always capture interpretable, concept-level properties of visual scenes presented to the agents. Rather than that, agents can learn specifically to successfully distinguish images while overfitting to inscrutable low-level relational properties of images. Agents can still solve guessing games with high accuracy with input data that contain no conceptual content at all, for example, when images are just random noise sampled from a Gaussian distribution \citep{bouchacourt2018}.

\subsection{Multi-task learning: simultaneous classification and signaling}

The structure of the emerging communication protocol may also be the result of demands to emit signals while simultaneously performing another action. This can be reflected by agent architectures capable of multi-task learning. This setup was explored by \cite{cangelosi1999, cangelosi2001}, who constructed agents foraging mushrooms of different types. Agents were rewarded for correctly identifying a mushroom type based on unreliable perceptual input and a signal produced by their peers. Each agent was controlled by its individual neural network. Networks were evolved through a simple genetic algorithm (population learning) and additionally fine-tuned using backpropagation (individual learning). The agent network had separate output layers responsible for identifying mushrooms and producing signals, and a hidden layer shared between these tasks. Communication had a structure similar to the basic signaling game, where an agent foraging for mushrooms was the receiver, and its parent from the previous genetic algorithm generation was the sender. The sender was presented with a true vector of perceptual properties of a given mushroom and produced a two-component signal. The receiver was presented with a noisy version of the vector of mushroom properties and with the sender's signal. It had a double task of correctly identifying the mushroom (rewarded in the evolutionary timescale) and recreating the sender's signal (trained with backpropagation in a supervised manner). This elaborate setup led to the emergence of a systematic communication protocol with disentangled mushroom properties.

To check the generalisation capabilities of these systems, Cangelosi applied a particular test based on \cite{deacon1997} interpretation of the classic experiments of teaching language to chimpanzees \citep{savage-rumbaugh1978, savage-rumbaugh1980}. The reasoning is as follows: let us assume that we know some basic vocabulary in one language. Then some new foreign vocabulary is introduced in a grounded manner, albeit for a limited number of examples. Next, the learner is presented with combinations of words in the foreign language that allow them to recognise a familiar grammatical rule. Afterwards, the learner should be able to correctly use the foreign words in combinations, even if they were not introduced explicitly for all examples. This schema was replicated 1-1 with the neural networks. The trained network represented the properties of mushrooms A and B with independent symbols. Cangelosi retrained the network using backpropagation to substitute existing symbols for property A for 2/3 of mushrooms with new symbols. Then, he introduced new symbols for property B and taught the network how symbols for properties A and B are composed together in two-component signals. Finally, he introduced new symbols for property A for the remaining 1/3 of the mushrooms. At this point, the network was able to correctly associate all symbols A and B, although this association was explicitly given only for 2/3 of mushrooms.

This kind of generalisation goes beyond generalisation of responses to new stimuli; this is a systematic transfer of relations formed between one group of symbols to a new group (in accordance with the Deacon semiotic framework). In terms of \cite{barrett2017} this could be called an example of template transfer, where the stimuli remain the same, but the response repertoire changes. \cite{cangelosi2002} considered this an example of “symbolic theft”, where preexisting grounded symbols and relations between them are used to ground new ones. Agents were able to learn and transfer this structure of relations because of their simultaneous participation in the production of signals and the classification of mushrooms.\footnote{In Cangelosi’s simulations agents were also moving in a 2D environment and had to approach a mushroom before picking it up. However, since they did not have any form of memory, their movement trajectory did not influence their classifications and did not contribute to the structure of their communication.}

\subsection{Modeling an ecological niche for communication}

\cite{loula2010} decided that to plausibly model phenomena related to language emergence, they need to broaden the scope of the model and simulate ecological relationships shaping a niche in which the communication occurs. They constructed an artificial life simulation inspired by the ecology of vervet monkeys. In the simulation, a bunch of herbivore creatures explored a 2D environment looking for edible plants while avoiding predators. Contrary to the basic LSG, signaling behaviours existed within the natural ecology of the simulated environment, going beyond static sender-receiver interactions.

Creatures in the simulation had capabilities to see nearby objects, hear vocalisations, and select predefined actions based on their perception and internal drives (hunger, fear, boredom, etc.). Actions were high-level behaviours, such as scanning the environment, wandering, fleeing, hiding, climbing a tree, vocalising, etc. The communication was not limited to sender-receiver pairs: any creature within vicinity was able to hear a vocalisation and react to it (possibly with another vocalisation). No particular reaction was distinguished as “correct”, systematic associations between vocalisations (signals) and predator types (meanings) spontaneously self-organised through Hebbian learning of associations between stimuli in this ecology. The authors demonstrated how the emergence of signals transforms ecology. At first, creatures reacted to signals by scanning their environment looking for potential predators: all signals were interpreted as general alarm calls signaling danger. Later, when signals were well learnt, the creatures took an action adequate for the predator type (fleeing, hiding, or climbing a tree) immediately after hearing the signal without ever seeing the predator. This marks a shift between different forms of signification, where the same signals gradually change meaning. This could be interpreted as another example of template transfer \citep{barrett2017}. As these simulations demonstrate, even a very simple association learning mechanism can stabilise communication if the ecology is rich enough.

\subsection{Embodied and situated communication}

In the vanilla LSG all actions are discrete: the model abstracts from action dynamics. \cite{mordatch2018} started from the standard sender-receiver interaction, but provided a minimum embodiment of their agents by placing them in a continuous environment where action and communication dynamics matter. In their simulation, a 2D continuous space was populated by agents and coloured landmarks. Agents learnt through supervised learning to communicate to each other about actions that they must undertake (e.g., go to) and objects that actions must target (e.g. blue). The communication protocol that emerged in throught the simulations was efficient and had a compositional verb-noun structure. Compositionality was encouraged through a soft limit on vocabulary size that made agents reuse the already known words (introducing a new word generated cost in the form of penalty). The ability of an agent to move freely contrasts with dominant settings, where the sender passively observes a stimulus (independently of any actions it can undertake) and the receiver is waiting for a message to undertake an action. The fact that the sender could move enabled nonverbal communication as the receiver used, for example, gaze direction or physical movement of the sender to inform its actions. Moreover, according to the authors, the physical properties of the environment and agent communication (e.g. actions and utterances that take time to accomplish) shaped the syntactic structure of the utterances. In the emerged communication protocol, the GOTO action type symbol was uttered before the landmark symbol, allowing the receiver to “prepare” for further actions by moving to an advantageous central position among the landmarks before starting to move towards the specific one.

Respect for the rich and structured environment as a source of structure appearing in language is clear in the experiments of \cite{floreano2007}, who go further with the embodiment of agents by employing mobile robots. The robots navigated through a 2D space to discover the food source and avoid the poison source. The robots used neural controllers evolved through an evolutionary algorithm, where fitness was determined by their food intake. They had an ability to emit light signals, helping other robots locate food, but this created competition for resources: spatial limitation of the environments allowed only a limited number of robots to feed at the same time. Contrary to the standard LSG, there were no sender-receiver roles, no turn-taking, no explicit reward for successful communication. The results indicated that cooperative communication can emerge among kin groups even if signaling is costly for individuals. More varied behaviours, such as the emergence of deceptive communication in groups of nonrelated robots, were also observed. Importantly, the values and costs of communication in these experiments were not arbitrarily introduced, but emerged from the properties of the environment.

\subsection{Combining concepts across games}

Yet another mechanism that leads to more complex communication was explored by \cite{schulz2012}, who experimented with mobile robots that navigated a natural environment (office space). They used language games similar to Steel’s guessing game, with different roles of sender and receiver. There was an explicit acquisition of new vocabulary by the receiver through imitation. The new element was that the agents played a couple of different games and the concepts acquired through these games were interrelated.

The robots had the ability to move freely in their environment and construct private maps of spatial experiences based on visual stimuli and self-motion odometry (using the RatSLAM algorithm \citep{milford2010}). When two robots met, they were able to hear unique signals that they emitted and thus establish that they are in the same location. They started by playing a \emph{where-are-we} game in which they named the location of their meeting. After a few games, the agents had conventionalised the lexicons of location names grounded in their private spatial maps. The robots then used grounded location names to play two new games: \emph{how-far} and \emph{what-direction}. In the how-far game two locations were named and robots had to negotiate words to express the distance between them, while in the what-direction game three locations were named and robots coordinated terms to express an angle between the two locations from the point of view of the third. After they coordinated their lexicons, they were able to use both location names and relational vocabulary to successfully play \emph{go-to} games when they arranged to meet at a specific location. Then, in \emph{where-is-there game} the agents used the relational vocabulary to name the locations they had not yet visited. The labels established during this last game remained abstract until the agents were able to discover previously unknown locations through exploration.

It is worth stressing that two different kinds of grounding of the vocabulary were present in this model. The first was a direct grounding in the experiences of spatial locations. The second was an indirect grounding in the relations between already established concepts. This can be considered as an example of building abstractions.

The communication complexity in this model is built incrementally starting from simpler games, and the games are combined in a modular composition schema \citep{barrett2017}. This requires a progressive complexification of agent interactions (here encoded in game structures) and a memory mechanism linking different concepts and allowing meaningful operations on them (here spatial maps of the environment).

\subsection{Situated coordinative dialogue}

The structure of the LSG brings to mind a descriptive function of communication: the sender describes the state of affairs unknown to the receiver. In contrast, \cite{grouchy2016} modelled an embodied and situated scenario where there are no external states to communicate, only agents that coordinate their movements in space to meet each other. They simulated colonies of mobile robots that navigated continuous 2D spaces using controllers evolved through genetic programming (evolved controllers were transferred to physical robots: e-pucks). The controllers took the form of ordinary differential equations that defined the relationship between robot sensors and actuators. The robots could not see each other but could hear continuous signals emitted by nearby peers. Whenever two robots met in space, they crossed their genetic material and produced an offspring. There was no explicit reward or fitness function, but the reproduction mechanism favoured robots that were able to find each other more efficiently.

Most populations of robots evolved communication systems in which the emitted signals corresponded to the robot's position in one of the two dimensions. Some populations used this communication strategy for a while but later on evolved more complex strategies, where signals emitted by two robots became codependent: a signal emitted by one robot was modulated by the signal emitted by the other robot. The established dialogical communication between the two robots became a dynamical control structure facilitating their meeting. The two modes of communication are dependent on each other in the sense that dialogical communication strategy emerges out of the simpler one. Conceptually, this resembles modular composition of signaling games proposed by \cite{barrett2017}, although here the signaling structure is fully emergent, without explicit roles, rewards, etc.

Grouchy's et al. model remains very original as it challenges many assumptions common in communication emergence models: a) it uses continuous signals instead of discrete signs, b) it does not reward agents explicitly, and c) it explores a form of communication that is less about representing states of the world and more about the dynamic coordination of behaviour (with respect to b) and c), \cite{floreano2007} model is also similar).

\section{Emergence, function of communication, and the mapping metaphor}

Reflection on the structure of the existing models demonstrates noble efforts to investigate various aspects of the communication emergence problem, yet at the same time reveals several gross categories of simplifications, which -- while useful in addressing concrete engineering problems or even concrete isolated theoretical questions -- might obscure the essential properties of human communication that need to be addressed in a successful explanatory model:
\begin{enumerate}
    \item Lack of emergence. In existing models, communication rarely truly emerges: symbols are predefined, given as discrete numbers or vectors. They usually form a separate modality, distinct from the perceptual input, which is already designed for communication by the model structure. The problems of conventionalization of replicable forms, tokenization of complex utterances, etc. are left out. The existence of discrete symbols is presupposed rather than explained. Agents’ actions also tend to be discrete and devoid of dynamics, preventing the emergence of new forms of organisation. This is somewhat relaxed in only a couple of models (pointing and “gaze” direction in \cite{mordatch2018}, movement and use of light properties for communication in \cite{floreano2007}, continuous signals and movement in \cite{grouchy2016}).
    \item Limiting the function of communication. In most of the existing models, communication has a single function which is explicitly rewarded. The focus on the descriptive function of communication is evident in the settings where there is knowledge asymmetry between agents, and one agent is rewarded for passing certain information about the world to the other. The imperative or coordinative function of communication, concerning the regulation of the ongoing interaction (for example, by eliciting certain behaviours of the interaction partner), is subordinate to the descriptive function (as in \cite{schulz2012}, where robots first learn descriptive communication and then use it imperatively). A notable exception is the model of \cite{grouchy2016}, which demonstrates how coordination can be independent of the description of the external environment.
    \item Mapping metaphor. The meaning of language is reduced to mapping discrete signals to discrete referents. This is visible in the basic LSG, in which the reward is maximised when the emerging signals can be uniquely mapped to the states of the world. Compositionality is understood in terms of more complex structured mappings \citep{korbak2020}. In the simplified environments adopted by most simulations, actions performed by agents and action history do not really contribute to the meaning of communication. An effort to break free from the mapping metaphor is visible in \cite{loula2010}, where emerging meaning is understood as a transformation of the ecological niche rather than simple mapping. Also in \cite{grouchy2016} signals allow functional coordination without mapping to specific referents.
\end{enumerate}

All these simplifications are in line with the basic structure of LSG as seen through the lenses of Shannon’s information transmission \citep{skyrms2010}. Although some existing models depart from these simplifications (usually not all three of them at once), in general, they hold surprisingly strong. We assume that this is due to a widespread approach to symbols, which treats them as something “given”, which exists independently of meaning, similar to unbound variables in a computer program ready to be associated with some values. Symbols are thus perceived as discrete abstract entities with formal properties that can be manipulated. Years of dominance of the information processing and computer metaphor in the cognitive sciences result in thinking about symbols as necessary assumptions prior to constructing explanations. This makes it easy to forget that their existence also needs to be explained. Consequently, the explanatory thrust is towards how we can manipulate symbols, combine them, and decipher their meaning. But the problem of the emergence of complex communication essentially includes the problem of symbol emergence: how do they come into being in the first place? What kinds of meaningful relations do they require to be described as symbols \citep{raczaszek-leonardi2018b}?

Assuming abstract symbols with arbitrary meanings as a starting point is also limiting because it puts the burden of structuring communication on the conventions created from scratch by the agents and realised by their internal architecture, neglecting or limiting the role that the environment and interactions with it perform. In the real world, the sources of forces that shape and structure communication are multiple. For instance, physical properties of the signals also take part in determining their meaning (like the light direction in \cite{floreano2007}), situated agents can only communicate effectively within some distance (as in \cite{loula2010}), etc. Neglecting the ways symbols might be created in interactions within a particular ecology obscures these sources of complexity, leaving us only with the simplified view of communication meaning as mapping from symbols to features \citep{raczaszek-leonardi2018c}.

\section{Sources of communication complexity}

Following the above review and the summary of theoretical considerations, we identify some of the most important sources of communication complexity and formulate the desiderata for a successful model that includes them.

\paragraph{Perceptual grounding}
Grounding language in perception distinguishes most of the emergent communication models from distributional semantic approaches, where language meaning is derived only from word co-occurrences. In perceptually grounded models, the agent's perception affects the language structure. For example, in \cite{steels2005} vocabulary structure depends on the size of just noticeable difference in agent’s perception. Models aiming to explore aspects of perceptual grounding should aim for realistic stimuli (for instance, pictures represented as raw RGB arrays as in \cite{lazaridou2018a}), but also realistic perceptual systems including multimodality, inherent limitations, perceptual noise, etc.

\paragraph{Sensorimotor grounding}
Another level is grounding language not only in passive perception, but in experiences gained from active exploration of the environment (active perception, \cite{gibson1966}). This requires an embodied agent capable of nontrivial actions that affect states of the world that later affect the perception of the agent (closing the sensorimotor loop). Simple extensions of the signaling game model or machine learning models inspired by image captioning or question answering tasks preclude that. Sensorimotor grounding is possible in embodied simulations, such as \cite{mordatch2018}, where agents move in the environment and perform sequences of actions, which in turn opens up additional possibilities for language complexification.

\paragraph{Properties of sign vehicles}
Communication complexity can also be induced by the physical properties of sign vehicles (signals) used in communication. As we observed before, many models opt for discrete and amodal signs and do not explore this dimension of communication complexity at all. In embodied models where signals have physical vehicles, such as the model of \cite{floreano2007} where light signals are used, the properties of vehicles and the properties of the perceptual systems of agents interact. This creates new constraints on the interaction scenarios, for example communication being possible only in vicinity. In addition, physical sign vehicles allow us to explore aspects of iconicity—similarity between a sign vehicle and its referent (be it an object, action, event, etc.)—which cannot be studied with amodal signs.

\paragraph{Complexity of the environment}
It is generally recognised that, in order for complex communication to emerge, a sufficiently complex environment is needed. Within the basic signaling game schema, degrees of freedom of signals are limited by the degrees of freedom of the stimuli. Simulations which deal with natural images as stimuli explore this dimension of complexity, although as \cite{bouchacourt2018} demonstrated using complex stimuli does not automatically mean that agents will arrive at meaningful conceptual representations. Another dimension of environment complexity, which appears in more embodied simulations, is the inherent environment dynamics: how does the state and/or changes in the the environment affect the agents and how, in turn, they are affected by it. All simulations that incorporate agent movement tap into this dimension. Finally, the environment may be fully observable—when the agent has full access to the environment state—or only partially observable. Even embodied simulations in the style of \cite{mordatch2018} tend to prefer fully observable environments, which are easier to learn by agents and in which communication is less ambiguous. The ambiguity introduced by partially observable environments may add to the complexity of communication. 

\paragraph{Agent architecture}
Internal agent’s architecture governs the available learning mechanisms. Humans are said to exhibit certain biases related, among others, to the limitations of working memory \citep{baddeley2003}. As we know from the No-Free-Lunch Theorem \citep{wolpert1996}, every learning algorithm has a preference towards a certain class of problems and thus necessarily introduces a bias favouring certain communication protocols over others. The strength of this bias is perhaps moderated by the flexibility of the algorithm. A simple neural network without a hidden layer used in the work of \cite{smith2003} could only represent the additively linear influence of object properties on signal components, while a multilayer network used in more recent studies could represent a wider class of relations \citep{ren2020}.

The scope of agent behaviour governed by learning differs between models. In the works of \cite{mordatch2018} or \cite{grouchy2016} an end-to-end learning is used, where all aspects of behaviour are governed by the same mechanism. In contrast, \cite{schulz2012} and \cite{loula2010} use specialised vocabulary learning mechanisms, representing other aspects of agent activity through preprogrammed procedures. On the one hand, these preprogrammed procedures structure the agent's interaction and, therefore, are a scaffolding and a source of structure for communication. On the other hand, the separation of agent activity into modules may preclude deeper interactions between communication, action, and perception.

There are also different design decisions in the implementations of learning algorithm depending on the notion of what is realistic in a simulation. For example, \cite{mordatch2018} use direct weight sharing between agents and full knowledge of the dynamics of the environment in the optimisation procedure that represents learning. This makes agents learning efficient but raises the question whether it does not circumvent some important challenges in language learning and miss additional sources of structure.

\paragraph{Communication function, agent goals and autonomy}
Obviously, communication can only evolve if it is useful in some sense. It does not seem that there is a single universal use of communication in the natural world. Communication appears in different forms, both when communicating individuals have a common goal and when the goals are different, in the context of cooperation between individuals and competition between them, both as faithful communication and as deceptive \citep{wacewicz2018}. In computational simulations, all these aspects are usually modelled as the fitness/cost function. What is included explicitly in the cost function is a nontrivial question. In the basic signaling game, successful communication is the goal of its own and is rewarded directly and symmetrically, while in \cite{cangelosi2001} or \cite{floreano2007} it is only a means to successfully locate the food and the reward is different for the speaker and the receiver. Most models consider cooperative scenarios, although some explore the conditions under which deceptive communication appears \citep{mitri2009}. In \cite{grouchy2016} there is no explicit cost function at all. Instead, artificial agents are allowed to reproduce whenever two of them meet in the same place (analogue of sexual reproduction); communication arises from evolutionary pressure. We argue that agents in this kind of simulation have more autonomy and a higher chance of evolving a communication system beyond the expectations of the modeler. On the other side of the spectrum there are simulations in which aspects of the communication protocol are directly included in the cost function, such as the pressure to keep the vocabulary small in \cite{mordatch2018}.

\paragraph{Social adaptations and social relationships}
Somewhat related to the above, pressures on the structure of communication arise from living in a society. Social pressures may come in two varieties: as special adaptations of individuals to live in a society, or as social scaffolding shaping the interactions of individuals. Adaptations for social behaviour in the form of mechanisms loosely inspired by the Theory of Mind are included in some models of emergent communication \citep{batali1998, jaques2019, bard2020}. They are reflected in the agent's internal architecture and affect the evolved communication systems. Then, social structures in the form of social networks \citep{zubek2017a}, social ecology \citep{loula2010} or kinship structures \citep{floreano2007, mitri2009} are shown to be important for the shape of emerging signals. Still, many works relying on the basic signaling game are limited to a one-sided communication between a pair of agents and do not study any such aspects of sociality.

\paragraph{History of interactions}
As with all complex evolving phenomena, natural languages bear the mark of their long history of evolution. The current shape of the communication system may be caused by the particular historical trajectory of the changing conditions that shaped it. On the level of individual language learning, there is a historical trajectory of language development that goes through different stages. Historicity understood in this way is explored only by a couple of models of emergent communication. The paradigm of iterated learning, which models vertical language transmission, taps into some aspects of historicity and the information bottleneck associated with it. In \cite{schulz2012} language games were played in a certain order, and the terms grounded using previous games were used in subsequent games. \cite{zubek2017a} studied reactions of a population of agents to a certain trajectory of changes in the environment and demonstrated that the structure of language categories is affected by history. \cite{korbak2021a} presented a developmentally motivated model in which an agent goes through two stages of learning, which helps it develop a compositional language. 

\section{Embodied models combine multiple sources of complexity}

Regarding different sources of communication complexity, we may treat them as semi-independent. It is reasonable that some models explicitly focus on a single source and attempt to isolate it from the others. This makes the model easier to interpret and helps to demonstrate the importance of the particular source. For example, \cite{steels2005} study perceptual grounding with minimal agents in an environment limited to colour stimuli. However, this choice should be conscious and well motivated and it should be recognised that the sources of complexity may be hard to isolate, as they can be easily conflated when sufficient complexity is introduced. Models which take seriously the notions of embodiment and situatedness operate under the assumptions which are of consequence for most sources of communication complexity listed above.

For instance, in \cite{floreano2007} physical space constraints (inability of multiple robots to occupy the same spot) had consequences for the socio-ecological function of communication. In this work, the robots had to locate and attach themselves to a food source, but because of the limited space only a certain number of robots could feed simultaneously. Thus, communicating the location of food posed the risk that some other agent would exploit the scarce resource. This implicit cost of communication, depending on the evolutionary selection mechanisms, resulted in various communication strategies, from honest signaling to deceptive communication \citep{mitri2009}). In some cases, the robots also learnt to exploit the physical properties of the medium of communication: they hid emitted light signals from the other robots by hiding behind the food source \citep{lehman2018}. This is an example of an emergent behaviour, which could not evolve in a simpler environment.

The environment adopted by \cite{grouchy2016} was minimalistic in the sense that it is devoid of any objects apart from the agents. Still, the agents were situated in the sense that they could freely move in continuous space and use their own locations as a context for communication. Importantly, the emitted signals were continuous and were interpreted as such. This opened new possibilities for meaning emergence, as the signal intensity could convey information on the distance in space (basic iconic form of significance). What emerged was not a static signal-referent mapping, which can be used to interpret only previously encountered signals, but a dynamical structure using signals to control agent actions that was capable of generalisation. Ultimately, that allowed for the dialogical form of communication where the signals of both agents became entangled to emerge.

Situatedness is also naturally connected with aspects of perceptual and sensorimotor grounding. It was shown that situating agents in more complex environments has a positive influence on systematic generalisation. Although not directly addressing the problem of language emergence, \cite{hill2020} presented a model in which they studied the effects of embodiment on language comprehension. Their agent had to navigate through the environment and perform actions according to the instructions given as a sentence in natural language. They demonstrated that (i) egocentric perspective (as opposed to third-person view), (ii) 3d simulated environments (as opposed to 2d grid worlds) and (iii) active perception (as opposed to perceiving the stimuli statically) incentivise the agent to factorise experience and behaviours into pragmatically motivated reusable chunks, leading to better generalisation as shown in experiments isolating factors (i)-(iii). These results provide an explanation of previous results, showing a low systematic generalisation in recurrent neural networks trained on symbolically encoded stimuli from a single modality \citep{lake2017}. \cite{hill2020} hypothesise that this effect arises because richer experience is a form of implicit data augmentation. This suggests that the human capacity to exploit the structure of the world, when learning to generalise in systematic ways, could be replicated in artificial neural networks if those networks are given access to a rich, interactive, multimodal stream of stimuli that better matches the experience of an embodied human learner.

\section{Conclusion}

Models of communication emergence come from multiple traditions, which explains the differences in terminology and emphasis. The lack of systematic understanding of the assumptions behind various models contributes to a barrier preventing researchers from one domain from building on the modelling efforts from another domain. We hope that our review will be helpful in building such a systematic understanding.

As is made clear by our analysis, the pressures shaping communication signs and symbols are many, and building a model means bracketing some of them out. Sometimes, however, the bracketing is too heavy and too hasty, as it may remove the vital elements of the explanation. Perhaps the fascination with one particular kind of a language game, involving a sender, a receiver, and clearly distinguished referents, made us forget that games people play are multiple \citep{berne1964}. We propose a change of perspective to the dynamic one advocating embodied and situated models, which naturally accommodate multiple forces, make room for recognising the historicity of communication systems, and do not presuppose the existence of readily distinguishable symbols. This requires freeing language from its sole role of world-describer to a broader function of interaction controller. Under such a view, the mapping metaphor is no longer adequate to explain the meaning of language and cannot be used as a foundation for communication emergence models. Instead, we propose to pay more attention to models where meaning emerges as a transformation of agents interaction in an ecological niche.

\section*{Declarations}

This work was funded by National Science Centre, Poland (OPUS 15 grant, 2018/29/B/HS1/00884). The funding source had no role in data collection and manuscript preparation. The authors declare no competing interests.

Authors' contributions: JZ, TK, JRL contributed to the conception, writing and editing of the manuscript.

\bibliography{communication_emergence}

\end{document}